\documentclass[11pt]{article}

\usepackage[preprint]{acl}

\usepackage{times}
\usepackage{latexsym}
\usepackage[T1]{fontenc}
\usepackage[utf8]{inputenc}
\usepackage{microtype}
\usepackage{inconsolata}
\usepackage{graphicx}
\usepackage{xkcdcolors}
\usepackage{amsmath}

\usepackage{booktabs}
\usepackage{fontawesome5}
\usepackage{multirow}
\usepackage{colortbl}
\usepackage{comment}
\usepackage{subcaption}
\usepackage{mathtools}
\usepackage{MnSymbol}
\usepackage{tabularx}
\usepackage{colortbl}
\usepackage{xcolor}
\usepackage{pgfmath}
\usepackage{array}

\usepackage{algorithm}
\usepackage{algorithmic}
\usepackage{amsthm}
\theoremstyle{definition}
\newtheorem{theorem}{Theorem}
\usepackage{newfloat}
\usepackage{listings}
\DeclareCaptionStyle{ruled}{labelfont=normalfont,labelsep=colon,strut=off} 
\floatstyle{ruled}
\newfloat{listing}{tb}{lst}{}
\floatname{listing}{Listing}

\title{\textit{Where is the multimodal goal post?} On the Ability of Foundation Models to Recognize Highlights in Videos}

\author{Aditya K Surikuchi, Raquel Fern{\'a}ndez, Sandro Pezzelle\\
  Institute for Logic, Language and Computation\\
  University of Amsterdam\\
\texttt{\{a.k.surikuchi|raquel.fernandez|s.pezzelle\}@uva.nl}}

\begin{document}
\maketitle
\begin{abstract}
Foundation models are used for many real-world applications involving language understanding and generation from temporally-ordered multimodal events. In this work, we study the ability of models to identify the most important sub-events in a video, which is a fundamental prerequisite for summarizing, reasoning over, or narrating multimodal events. Specifically, we focus on football games and evaluate models on their ability to distinguish between important and non-important sub-events in a game. To this end, we construct a new dataset by leveraging publicly available football highlight reels containing the most important moments of a game according to professional experts, without any additional annotation costs. Using our dataset, we compare several state-of-the-art multimodal models and show that they fail to reliably capture contextual importance. Analyses of models beyond standard evaluation metrics reveal their tendency to rely on a single dominant modality and their ineffectiveness in synthesizing necessary information from multiple sources. Our findings underline the importance of modular architectures that dynamically adapt to multimodal heterogeneity and the need for complementary training procedures that maximize cross-modal synergy.
\end{abstract}

\section{Introduction}
\label{sec:1}
Recent advancements in artificial intelligence have facilitated the development of models, referred to as `foundation', `omni', or `multimodal' models, that can process several modes of data, including sequences of images, videos, and audio, to automatically generate descriptions, summaries, or stories about temporally-ordered multimodal events in natural language \cite{unified_io,qwen2_5_omni,qwen3_omni}. These capabilities, along with the increase in availability of large datasets, have propelled their widespread adoption for groundbreaking applications such as real-time automatic commentary generation in sports \cite{MatchTime,SR-1988,TimeSoccer}. Despite this impressive progress, the practical applicability of these models remains an open question. While the outputs generated by these models may seem fluent and plausible at first glance, assessing their quality in terms of visual grounding, coherence, topic consistency, \textit{inter alia}, is challenging \cite{assess_mm_skills,NYTWS,narratives_from_VE,assess_mm_gen}. Furthermore, there is increasing research uncovering the limitations of multimodal models, including their difficulty with counting objects, temporally tracking entities, and reasoning spatially \cite{spatial_reasoning,vilma,temporal_bias,entity_tracking}.

\begin{figure*}[htbp]
  \centering
  \includegraphics[width=0.94\textwidth]{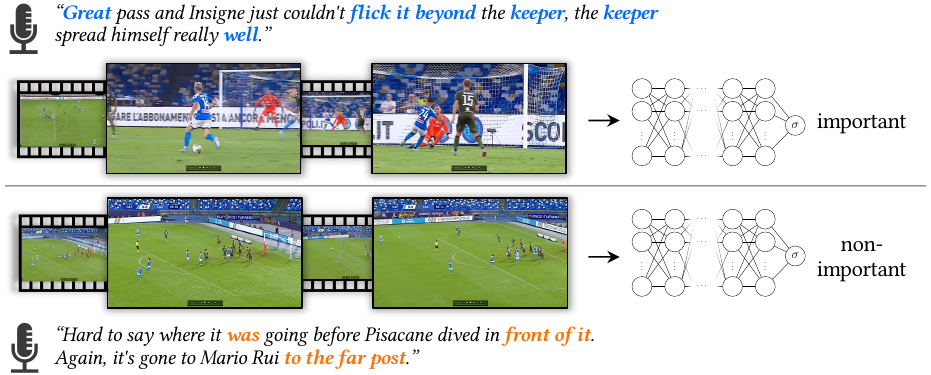}
  \caption{The clip on the top, showing a shot-on-target, was featured in the highlights of the football game to which the clip belongs. In contrast, the one on the bottom, showing a corner kick, was not. A logistic regression classifier trained on the transcriptions of commentaries identifies the text in {\color{xkcdCeruleanBlue}\textbf{blue}} as driving the prediction toward the `important' class, and the text in {\color{xkcdOrange}\textbf{orange}} as driving it toward `non-important'. The confidence of a foundation model, Qwen-2.5-Omni, in correctly identifying this clip featured in the highlights reel as `important' is highest under vision-only input (\underline{V=3.81}, L=-0.18, LV=-0.93), whereas for the `non-important' clip multimodal information (language in the commentary complementing the visual) results in the highest model confidence (V=0.5, L=0.87, \underline{LV=1.34}). More examples from our dataset are provided in Figure~\ref{fig:moments_examples_supplementary} of Appendix~\ref{appendix:C}.}
  \label{fig:moments_examples}
\end{figure*}

In this work, we take a step back from the actual narration of visual events and focus on a fundamental prerequisite ability: \textit{identifying the most important sub-events in a (long) video}. This ability is essential for any generative task because a coherent and compelling narration of a multimodal event is primarily dependent on identifying and comprehending its key moments. To this end, we fill a gap in the literature by constructing a novel dataset to test models for this crucial ability. Specifically, we focus on football (\textit{soccer}) considering the open availability of full game videos, professionally narrated live commentary, and broadcaster-produced highlight reels. We create a new dataset that enables us to test, at scale and without additional annotation costs, whether models can effectively integrate commentary and information in the video to distinguish salient moments. In this context, we consider a moment in a football game as `important' if it is included in the highlight reels put together by experts in the broadcasting domain. Our dataset allows us to move beyond dimension-specific evaluations of generated text and conduct deeper analyses of model behavior and internal confidence (see Figure~\ref{fig:moments_examples}). We ask two main research questions: \textit{RQ1. To what extent can current state-of-the-art multimodal models distinguish between important and non-important moments in a football game?} \textit{RQ2. How effectively can these models leverage the natural multimodal signals in this data?} In particular, do models that receive multiple signals---video, audio commentary, and language transcription---outperform models that rely only on one of these modalities?

Given the predominantly visual nature of sports, for moments that are typically considered important in football, e.g., goals, we expect the visual modality (video) to encapsulate a stronger signal compared to corresponding commentaries or their transcriptions. However, we argue that the importance of most moments will be contingent upon several aspects---background information, tactical insights, strategic context, and the emotions evoked in audience and commentators---all of which extend beyond the limits of the visual modality. Consistently, we hypothesize that commentaries associated with these moments will be crucial to understanding their importance, based on the assumption that a professional commentator would highlight these aspects in their narration. Under this premise, we expect models that are effective at integrating multimodal information to perform better on the task compared to their unimodal counterparts.

Rather than measuring success solely by considering final model-generated outputs, we leverage model logits to examine cross-modal synergy. Through comprehensive experiments across multiple models (including Gemini \cite{gemini}, a proprietary model), modalities, and prompts, we demonstrate that current models are far from perfect at recognizing key moments included in game highlights, underscoring the inherent complexity of this task. While the closed model achieves the highest overall performance, its predictions still substantially diverge from the editorial ground-truths curated by professional experts. Moreover, we find that open-source multimodal models do not have a dramatic advantage over models that rely only on one modality (e.g., vision-only or language-only settings), demonstrating the ineffectiveness of current architectures in integrating and exploiting important multimodal information. Even for moments whose importance is contingent on aspects beyond the purely visual level (e.g., corners or shots on target), multimodal integration improves model confidence only in isolated cases, such as the examples in Figure~\ref{fig:moments_examples}. Among the three modalities, the visual modality has the strongest predictive power for correctly identifying important moments, confirming our expectations. However, for correctly recognizing non-important moments, we find that the commentator's information through natural language provides the strongest signal. This confirms our intuition that combining details from the visual modality and its associated commentary is essential for achieving high task performance.

Overall, our results show that current multimodal models do not accurately capture key moments in temporally-ordered events, indicating that substantial improvements in multimodal fusion and event understanding are needed before these models can be deployed for live narration of unfolding events or video summarization. Our dataset and code are available at: \faGithub\;\href{https://github.com/akskuchi/MOMENTS}{akskuchi/MOMENTS}

\section{Related Work}
\label{sec:2}
\subsection{Detecting Significant Moments in Videos}
\label{sec:2.1}
Automatically detecting significant moments from events unfolding over prolonged durations is a computer vision task that is crucial for several domains, such as advertising and sports. Most of the existing work formulates highlight detection as the ranking task of assigning a higher numerical score to interesting clips compared to non-interesting ones from the same video \cite{basketball_highlights, highlights_detection_ranking1,highlights_detection_ranking2}. Others define this problem as a binary classification task of identifying salient vs.~non-salient clips within a given input video \cite{highlights_detection_classification,cricket_highlights,dellasanta2025automateddetectionsporthighlights}. In both instances, the proposed approaches typically rely on training regression or classification models using some form of supervision signal such as user engagement ratings \cite{highlights_detection_ranking3} obtained through crowdsourcing or frame-level human annotations \cite{frame_level_annotations}. To bypass the costs of collecting importance labels from humans, recent work proposed an approach to obtain pseudo-highlight labels based on the notion of pattern recurrence \cite{unsupervised_highlights_detection}. Specifically, the method first organizes videos into groups using a clustering algorithm. It then assigns importance scores to clips within each video of a group based on its audio-visual feature recurrence, hypothesizing that significant moments, such as crowd cheering for a goal, tend to repeat across clips found in videos clustered into the same semantic group.

Contrary to all previous work that relies either on predefined heuristics or on an explicit human annotation process for determining the importance of moments, we leverage publicly available highlight reels of football games. Specifically, we consider moments that are included in the highlight reels created by experts in sports broadcasting as reliable proxies for important moments, given that the main emphasis of a highlights editorial process is to selectively filter the raw game into a coherent narrative comprising key moments \cite{proxy_for_importance}. This approach allows us to construct a high-quality multimodal dataset with a naturalistic distinction between important vs.~non-important moments.

\section{Constructing the MOMENTS Dataset}
\label{sec:3}
In this section, we describe the construction of our dataset, which we call MOMENTS. Our objective is to obtain a corpus of video fragments  (hence, \textit{moments}) extracted from whole football games, that are labeled as `important' or `non-important'. We intend to achieve this without following any predefined heuristics on what makes a moment important. Specifically, we propose a novel algorithm that uses visual information to \textit{localize} fragments of official highlight videos in corresponding full games for obtaining important moments. To this end, we use the GOAL dataset \cite{GOAL}, which contains highlight videos for football games from major leagues of Europe, between 2018 and 2020. Mapping highlights in GOAL with two publicly available full game video datasets, i.e., SoccerNet \cite{soccernet} and SoccerReplay-1988 \cite{SR-1988}, results in 127 <highlight video ($\textbf{H}$), full game video ($\textbf{G}$)> pairs.

\subsection{Localizing Highlights in Full Games}
\label{sec:3.1}
The goal of this stage is to find the closest visually matching frame from $\text{G}$ for every frame in $\text{H}$. This is based on the underlying assumption that $\text{H}$ is essentially a subsequence of $\text{G}$. Although this is theoretically the case, there are several caveats to consider in practice. Firstly, the auxiliary information in frames of $\text{H}$ is different from the information present in the exact same frames of the corresponding $\text{G}$ video. This information can include aspects such as advertisement overlays, score cards, and watermarks. Furthermore, $\text{H}$ videos typically contain opening and closing moments of the game, which are absent in the $\text{G}$ videos provided in SoccerReplay-1988. These aspects rule out the possibility of exact matching between frames of $\text{H}$ and frames of $\text{G}$. As a workaround, we rely on checking the similarity of two frames using multi-scale Structural Similarity Index Measure (SSIM) \cite{SSIM}, which compares the overall visual structure and composition (spatial patterns formed by neighboring pixels) between two frames and provides a similarity score in range [-1, 1], where a score of 1 indicates high visual similarity.\footnote{SSIM (\textit{dis})-similarity scores between frames from an <H,G> pair are provided in Figure~\ref{fig:SSIM} of Appendix~\ref{appendix:A}.}

On average, $\text{H}$ videos contain 6545 frames ($\approx$3 minutes) and $\text{G}$ videos contain 135000 frames ($\approx$90 minutes) making exhaustive all-frames comparison between $\text{H}$ and $\text{G}$ computationally prohibitive in practice, i.e., $\mathcal{O}($\text{H}$ \times $\text{G}$)$.\footnote{$\text{H}$ videos have 25 frames per second (FPS), and FPS for $\text{G}$ videos can range between 25 and 150.} We address this challenge by devising a three-step approach that hierarchically localizes $\text{H}$ in $\text{G}$.

\paragraph{Step 1.}
We first compare the initial frame of each second in $\text{H}$ with a representative frame (the center-most one) from every second in $\text{G}$. Then, we compare each initial frame in $\text{H}$ with all the frames \textit{only} at those seconds in $\text{G}$ where the corresponding representative frames are sufficiently similar (SSIM threshold of 0.8; chosen empirically) to the $\text{H}$ frame under consideration. This reduces computational complexity from $\mathcal{O}($\text{H}$ \times $\text{G}$)$ to $\mathcal{O}(\text{H}_{\text{1-FPS}} \times \text{G}_{\text{1-FPS}})$.

\paragraph{Step 2.}
Step 1 relies on approximate comparisons and may result in frames at some seconds of $\text{H}$ not being accurately localized in $\text{G}$. Specifically, these frames may have corresponding closest matching frames in $\text{G}$ with visual similarity below an acceptable predefined threshold. Therefore, as a subsequent step, we leverage $\text{G}$ frame boundaries of seconds in $\text{H}$ that are `well-localized' after step 1, to \textit{prune} the search space and improve localization for the remaining `poorly-localized' seconds of $\text{H}$. Among the 127 <$\text{H}$,$\text{G}$> pairs we process, more than half of the frames in 100 $\text{H}$ videos are accurately localized in their corresponding $\text{G}$ videos after step 2.\footnote{In Appendix~\ref{appendix:A} (Figure~\ref{fig:after_steps_1_2}), we plot the proportion of seconds in $\text{H}$ that are accurately localized after this step to show its importance. We also conduct human verification on a small subset of the extracted moments to check for sanity.}\looseness-1

\paragraph{Step 3.}
For seconds in H that are accurately localized in G after step 2 (in terms of their initial frame), we loop around the matched frame indices in G, to find matches for all the remaining frames of H. Finally, we group the resulting $\text{G}$ frame indices based on adjacency to obtain `important' moments. In this grouping, we use a separation threshold of 1 second for considering two moments as separate.

\subsection{Sampling Non-Important Moments}
\label{sec:3.2}
We consider contiguous segments of G videos that do not contain any `important' moments as `non-important'. To avoid any spurious patterns that may inadvertently reveal the importance of moments, we sample non-important moments with time durations similar to that of the localized `important' moments. Specifically, we fit a Gamma distribution to important moment durations, sampling from it the durations for non-important ones (see Figure~\ref{fig:moments_duration} in Appendix~\ref{appendix:A}).

\subsection{Extracting Moments from Game Videos}
\label{sec:3.3}
Finally, we convert the frame boundaries obtained for important and non-important moments into timestamps, and extract for each moment, the video frames and audio commentary from corresponding $\text{G}$ videos. To transcribe the audio commentaries of $\text{G}$ videos into text, we use the open-source \texttt{whisper-turbo} model \cite{whisper}. We also ensure that the clipped moments are `complete' in terms of their commentaries (i.e., not abruptly cut off mid-sentence) by leveraging as guidance the segment timestamps obtained as part of the $\text{G}$ audio transcriptions. Furthermore, to address the inherent latency that manifests as the \textit{eye--voice span} (EVS) \cite{EVS1,EVS2} in real-time naturalistic events, we extend the duration of auditory and textual modalities of each moment by three seconds beyond the rightmost timestamp of its video. We empirically verify the existence of this latency between the occurrence of a visual event and the commentator's narration of it. Overall, we extract 1977 important moments from the 100 most accurately localized <$\text{H}$,$\text{G}$> pairs, and to balance the dataset, we include the same number of non-important moments resulting in a total of 3954 samples (see Figure~\ref{fig:moments_duration} in Appendix~\ref{appendix:A} for duration comparisons).

\section{Experiments}
\label{sec:4}
In this section, we compare the performance of various models in their ability to distinguish between moments based on their importance. We operationalize this as a binary classification task using the proposed MOMENTS dataset.

\paragraph{Task Definition.}
Given a moment ($\textbf{M}_i$) from a football game, we task models in classifying it as \textit{important} or \textit{non-important}. Formally, we define the task as:
\begin{align}
  \Phi(\textbf{M}_i) &=
  \begin{cases}
    1, & \text{if } \textbf{M}_i \text{ is \textit{important}} \\
    0, & \text{otherwise}
  \end{cases}
\end{align}
where $\Phi$ refers to the model families we outline in Section~\ref{sec:4.1}. Every $\textbf{M}_i$ in MOMENTS is represented by three modalities---auditory (A), language (L), visual (V)---enabling comparison of compatible models under seven combinations of modalities $\mathcal{C} = \{\textbf{A}, \textbf{L}, \textbf{V}, \textbf{AL}, \textbf{AV}, \textbf{LV}, \textbf{ALV}\}$.\footnote{Where \textbf{AL} = \{A, L\}, \textbf{ALV} = \{A, L, V\}, etc.}

\subsection{Experimental Setup}
\label{sec:4.1}
\paragraph{Off-the-shelf pre-trained models.} We consider a diverse set of open-source pre-trained models from four categories and prompt them off-the-shelf for the task. Specifically, we experiment with three \textbf{language models}: Meta-Llama-3.1-8B-Instruct, Qwen2.5-7B-Instruct, and Qwen3-4B-Instruct; two \textbf{vision-language models}: Qwen2.5-VL-7B-Instruct and Qwen3-VL-4B-Instruct; two \textbf{audio-language models}: Qwen2-Audio-7B-Instruct and Voxtral-Mini-3B; and \textbf{two omni models}: Qwen2.5-Omni-7B and Qwen3-Omni-30B-A3B \cite{llama3,qwen2_5,qwen3,qwen2_5_VL,qwen3_VL,qwen2_AL,voxtral_AL,qwen2_5_omni,qwen3_omni}. In addition to these open-source models, we also test Gemini-2.5-Flash, a proprietary omni model \cite{gemini}. To check if models are sensitive to variations in the instruction, we assess them using the two prompts in Table~\ref{tab:prompts} (see Appendix~\ref{appendix:B}). Specifically, for each target class (`YES', `NO'), we average the prediction probabilities from models over the two prompts and classify a moment as important if the average probability of `YES' exceeds 0.5. Other inference details are provided in Appendix~\ref{appendix:B}.

Recent research has revealed that multimodal models are sensitive to the order in which modalities are sequenced within an input prompt, in that they respond differently to different orders \cite{order_confound1,order_confound2}. For instance, under the \textbf{ALV} combination, modalities in the prompt can be permuted in six different orders---ALV, AVL, LAV, LVA, VAL, VLA. To mitigate the computational overhead of exhaustively considering all possible permutations of modalities for all models, we employ a heuristic approach. We select one model per category, identify the `optimal' permutation of modalities that yields the highest score, and then test the remaining models in each category using this `optimal' permutation.

\paragraph{Baselines.} For each of the three modalities in scope, we learn a logistic regression classifier as a baseline by utilizing a 3:1 train:test split of MOMENTS. We use $n$-gram counts and Mel-Frequency Cepstral Coefficients (MFCCs) as features for textual transcriptions and audio waveforms respectively. We extract 1--4 grams for text and the top 20 MFCCs for representing the most important characteristics of the audio. To obtain features for the video modality, we extract 1 frame per second for each video, obtain SwinTransformer \cite{SwinTransformer} embeddings for the extracted frames, and consider the average of embeddings as the feature representation of the video.

We evaluate model responses using several metrics widely used for classification tasks including accuracy, F1 score, ROC AUC, and Matthews Correlation Coefficient \cite[MCC;][]{MCC}. However, in this section we primarily focus on MCC, considering that it accounts for all four aspects of the confusion matrices unlike other metrics such as F1 score that are biased towards true positives \cite{f1_shortcoming}. Given that our dataset is balanced in terms of the two classes, we also report accuracy (scores from other metrics are provided in Figure~\ref{fig:scores_supplementary} of Appendix~\ref{appendix:B}).

\begin{figure*}[tbp]
  \centering
  \includegraphics[width=\textwidth]{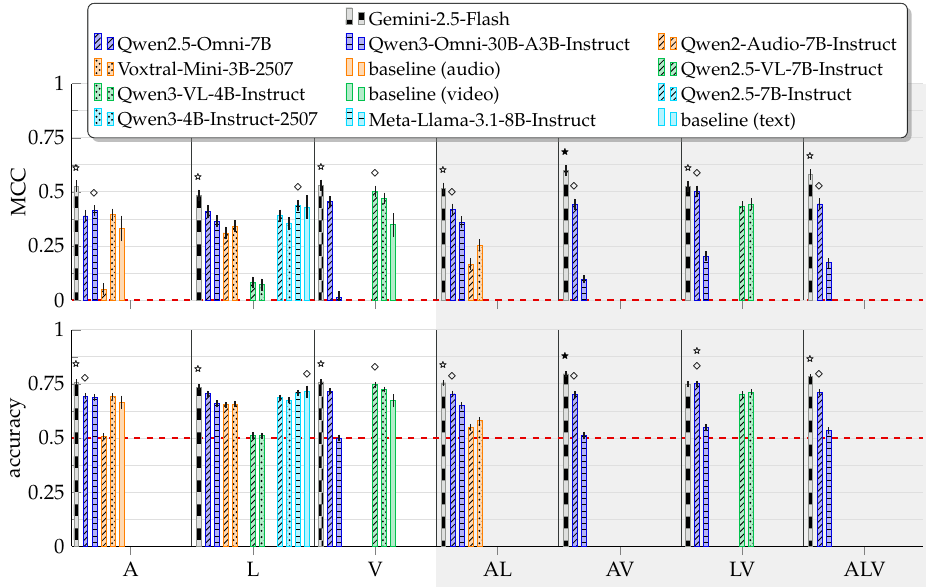}
  \caption{MCC and accuracy for all models under various combinations of modalities. We compute uncertainties (line indicators on each bar) representing the 95\% confidence interval, estimated via bootstrap resampling \cite{bootstrap_resampling}. Under each modality combination, $\smallstar$ denotes the highest scoring model and $\diamond$ denotes the highest scoring open-source model. $\filledstar$ represents the best performing model and combination of modalities.}
  \label{fig:scores_main}
\end{figure*}

\subsection{Results}
\label{sec:4.2}
Figure~\ref{fig:scores_main} shows the evaluation results we obtain for all the models under different combinations of modalities. Firstly, it is evident from the scores that the task is challenging and remains far from solved. The overall highest scores are obtained by a proprietary closed-source model (Gemini-2.5-Flash) under the \textbf{AV} combination of modalities; MCC = 0.598 $\pm$ 0.02, accuracy = 0.795 $\pm$ 0.01. While these results may indicate that models are able to extract signals from input data required for determining the importance of moments, their predictions are still substantially different from ground-truth editorial selections implicit in professional highlight reels (see Table~\ref{tab:confusion_matrices} in Appendix~\ref{appendix:B}).

Next, we find that open-source foundation models are not far behind the proprietary model we consider. Specifically, Qwen2.5-Omni-7B (under the \textbf{LV} combination of modalities; MCC = 0.502 $\pm$ 0.03, accuracy = 0.751 $\pm$ 0.01) and Qwen2.5-VL-7B-Instruct (under \textbf{V}; MCC = 0.502 $\pm$ 0.03, accuracy = 0.746 $\pm$ 0.01) obtain marginally better scores compared to all the remaining settings. However, we also observe that the performance across several open-source models and settings is largely comparable, with no model significantly outperforming others within the margin of uncertainty. Furthermore, these evaluation results do not reveal any clear statistically significant advantage for models receiving a multimodal input. In terms of the three modalities, we notice a slight performance difference between the \textbf{A} and \textbf{V} settings, where the overall best score under \textbf{V} is significantly better than the overall best under \textbf{A}. This trend however is not consistent for all models; Qwen3-Omni-30B-A3B-Instruct performs significantly better under the \textbf{A} setting compared to \textbf{V}.

Although these results are useful for reflecting on general surface-level performance, these metrics are not informative regarding the different ways models extract and integrate multimodal information. To complement these results and shed light on this issue, we analyze models' internal behavior.

\section{Analyses}
\label{sec:5}
\subsection{Influence of Modalities}
\label{sec:5.1}
In this analysis, we shift from an external evaluation of task performance (based on MCC and accuracy) to examining the internal models’ behavior, as reflected in their confidence scores. In particular, we propose a novel approach for quantifying the contribution that each modality makes to a model's confidence. For each moment, we compute the contribution of a modality ($m$) as the difference between the sums of logit differences of all the applicable sets of modality combinations ($\mathcal{C}$) that include and exclude $m$. Formally, we define
\begin{equation}
  \!\text{contribution}_m\!=\!\sum_{\forall c_i \in \mathcal{C}}\!{\Delta}Z_{m \in c_i}\!-\!\sum_{\forall c_i \in \mathcal{C}}\!{\Delta}Z_{m \notin c_i}
  \label{eq:2}
\end{equation}
where ${\Delta}Z$ is the difference between logits predicted for the ground truth class and the incorrect class. For instance, when considering Qwen-Omni models, determining the contribution of the visual modality (V) involves considering the difference between the sums of logit differences of four combinations in $\mathcal{C}$ that include V (\textbf{V}, \textbf{LV}, \textbf{AV}, \textbf{ALV}), and three combinations that exclude V (\textbf{A}, \textbf{L}, \textbf{AL}). Using this approach, we compute modality-level contribution scores over the entire dataset for all models, normalizing the score of each modality $m$ by the number of combinations containing $m$ so that scores are comparable across models.

\begin{figure}[tbp]
  \centering
  \begin{subfigure}{\columnwidth}
    \centering
    \includegraphics[width=0.88\columnwidth,keepaspectratio]{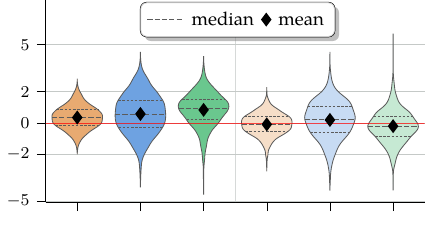}
    \caption{Qwen2.5-Omni-7B}
  \end{subfigure}
  \begin{subfigure}{\columnwidth}
    \centering
    \includegraphics[width=0.88\columnwidth,keepaspectratio]{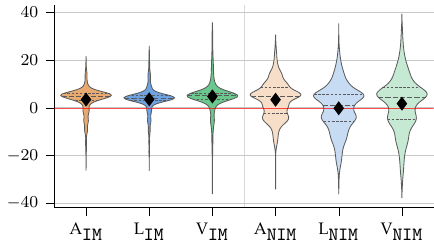}
    \caption{Gemini-2.5-Flash}
  \end{subfigure}
  \caption{Contribution scores of the three modalities for: (a) Qwen2.5-Omni-7B (b) Gemini-2.5-Flash, separated for important (\texttt{IM}) and non-important (\texttt{NIM}) moments.}
  \label{fig:contribution_scores_main}
\end{figure}

Figure~\ref{fig:contribution_scores_main} shows the contribution scores for the three modalities for the Qwen2.5-Omni-7B and Gemini-2.5-Flash models. Unlike open-source models, the Gemini API does not expose raw logits and returns only the top 20 log probabilities during inference. Consequently, the scores cannot be computed if a target class token falls outside this restricted window. However, in our experiments, we always retrieve both target classes within this window and can reliably compute the contribution scores. Furthermore, we formally show in Appendix~\ref{appendix:C} that the log-probability difference is mathematically equivalent to the logit difference (${\Delta}Z$). We observe that the visual modality contributes the most to the confidence of models for correctly identifying important moments of the game. However, the associated commentaries (audio for Gemini-2.5-Flash and textual transcriptions for Qwen2.5-Omni) play a more crucial role in determining the confidence of models for correctly classifying non-important moments. We find this pattern to be largely consistent across models that process video, whereas audio-language and unimodal models show no consistent pattern (see Appendix~\ref{appendix:C} for contribution scores of all considered models). This reveals that recognizing different kinds of moments requires reliance on different modalities to different extents. It also underlines how commentaries are crucial to complement and regularize information pertaining to event types that are key only in certain contexts. Indeed, two events of the same type (e.g., a free kick) may not both be important \citep{VAEP}, and this is likely to become apparent from the accompanying commentaries. Based on these insights, we postulate that models must rely on multimodal information beyond just the visual modality to recognize moments that are not prototypically important. To verify this, we study the role of multimodality for various types of moments.

\subsection{Role of Multimodality}
\label{sec:5.2}
To conduct this analysis, we identify 50 moments (from various games) where a goal is scored and consider them as prototypically important (indeed, we verify that they all belong to the class of `important' moments in our dataset). In addition, we find 100 moments pertaining to corner kicks, throw-ins, and shots-on-target, following the definitions provided by \citet{action_definitions}, which are important only in certain contexts. We find the moments of each type in two stages. First, we prompt Qwen2.5-Omni-7B to decide whether a moment shows the action, separately under each of the three modalities and record its confidence (${\Delta}Z$; prompts in Appendix~\ref{appendix:C}). Then, we keep moments that are classified reliably---${\Delta}Z > 3.0$ \cite{logit_difference_range}---as ``YES'' under all the three modalities, and verify them manually for correct type and complete action. Of the 100 contextual moments, 55 belong to the class of `important' moments, while 45 are `non-important' ones. This confirms our intuition that, for most of the moments in our dataset, excluding prototypically important ones, such as goals, there is no trivial one-to-one relationship between the type of the event and its importance (examples of selected moments are provided in Figure~\ref{fig:moments_examples_supplementary}, Appendix~\ref{appendix:C}).

\begin{figure}[tbp]
  \centering \includegraphics[width=0.95\columnwidth,keepaspectratio]{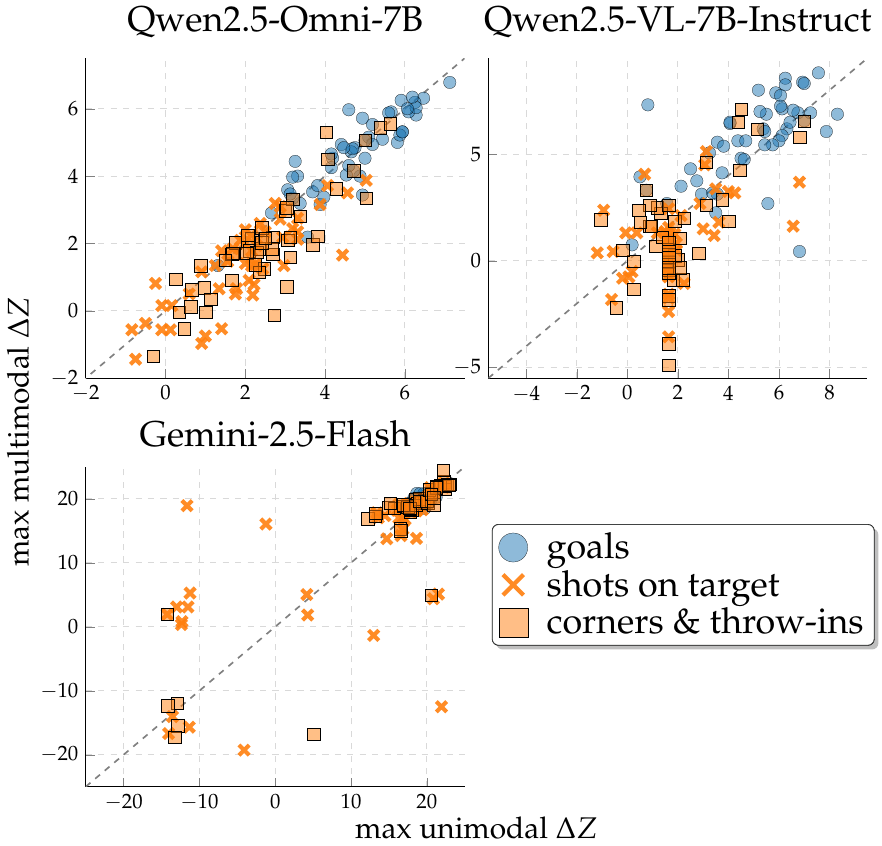}
  \caption{For each {\color{xkcdCeruleanBlue}\textbf{prototypical}} and {\color{xkcdPumpkinOrange}\textbf{contextual}} moment (important or non-important), we visualize model confidence (${\Delta}Z$) under optimal unimodal vs.\ multimodal settings. The diagonal ($y = x$) indicates equivalent confidence, while points on and below it signal unimodal dominance and cross-modal interference, respectively. Absence of points above the diagonal highlights limited multimodal synergy in open models.}
  \label{fig:multimodal_vs_unimodal}
\end{figure}

We assess the contribution of multimodal information by analyzing changes in the confidence of models (measured using ${\Delta}Z$) for the identified subset of moments. Specifically, for each moment in the subset, we map the optimal unimodal ${\Delta}Z$ against the optimal multimodal ${\Delta}Z$. Figure~\ref{fig:multimodal_vs_unimodal} visualizes this comparison for three models. Firstly, we observe a major difference between model classes, i.e., open-source models are more confident in determining the importance of prototypical moments (goals) compared to contextual ones. Specifically, goals (blue dots) cluster in the top-right regions of the plots, while corners, throw-ins, and shots-on-target are predominantly in the bottom-left regions. In contrast, the proprietary model exhibits high confidence across both categories of moments---most points are scattered in the top-right region.

The diagonal dashed line ($y = x$) in the plots represents the boundary where max multimodal ${\Delta}Z$ equals max unimodal ${\Delta}Z$. When a model integrates complementary information from multiple modalities for a given moment, we expect it to be positioned above this diagonal in the top-left region. Conversely, points lying on the diagonal signal unimodal dominance, whereas points below the diagonal in the bottom-right indicate cross-modal interference when combining modalities.

Ideally, if open models are truly multimodal, they should be capable of integrating information from multiple modalities to correctly recognize contextually important moments. However, this ideal is not reflected in actual results, as evidenced by the lack of many points in the top-left region above the diagonal. In the case of Qwen2.5-Omni-7B, we notice many points scattered below the diagonal, reflecting a collapse towards one of the modalities \cite{modality_collapse}. With Qwen2.5-VL-7B-Instruct, we observe a slightly more dispersed pattern, indicating that multimodality potentially introduces noise resulting in low confidence and misclassifications. In contrast, the proprietary model's ability to confidently classify contextual moments suggests that it is leveraging cross-modal synergy and mitigating modality collapse even for moments that are misclassified by open-source models, although its specific architectural details are undisclosed.\looseness-1

\section{Conclusions}
\label{sec:6}
In this work, we focused on the ability of models to recognize important moments in temporally-ordered multimodal events. We proposed a novel framework to construct a dataset of moments from football games by leveraging publicly available football highlight reels curated by professional experts. In contrast to concurrent work by \citet{video_summarization} which introduces a semi-automated pipeline for video summarization based on a similar intuition (considering moments in highlight reels as `important'), our framework prioritizes full automation and fine-grained localization. We achieved this by hierarchically matching all frames in highlight reels with frames at localized intervals in full game videos using a deterministic metric (SSIM) rather than latent feature backbones like DINOv2 \cite{dinov2}.

Assessing several state-of-the-art foundation models using our proposed dataset revealed their limited ability to distinguish between important and non-important moments in football games, as evidenced by evaluation scores in Section~\ref{sec:4}. Beyond downstream task metrics, we use prediction logits to quantify the contribution of each modality. Our analysis showed that models rely predominantly on vision for recognizing important moments and on text for non-important ones. This unimodal dominance suppresses models' ability to effectively integrate multimodal signals. Empirical assessment in Section~\ref{sec:5} exposed this limitation of open-source models, confirming that current multimodal architectures and training procedures are ineffective in disentangling synergies and redundancies from cross-modal information \cite{synergy_redundancy}.\looseness-1

This motivates the need for architectures that dynamically integrate multimodal information at the sample level, unlike existing architectures that are focused on static fusion strategies using the projector component \cite{narratives_from_VE}, which result in fixed information flow regardless of sample-level nuances, conflicts, and ambiguities between modalities \cite{sample_level_dynamics1,sample_level_dynamics2,sample_level_dynamics3}. A recent approach, Mod-Squad \cite{multimodal_MoE}, proposes a mixture-of-experts to route samples to modality-specific experts dynamically. We advocate for future work to design modular architectures with complementary training objectives that can account for the heterogeneity inherent in multimodal data.

\section*{Limitations}
We focus on football games considering the open availability of full game videos, live commentary, and broadcaster-produced highlight reels. While highlight reels put together by experts in the broadcasting domain provide reliable proxies for important moments without additional annotation costs, our findings may not directly extend beyond sports to other multimodal events where key sub-events lack such an editorial process.

Moreover, as a first step, we consider moments as, to some extent, independent from each other. Investigating how the importance of a moment is related to other moments in the game is an interesting direction for future work.

Furthermore, we evaluate models off-the-shelf, without fine-tuning or in-context examples, which might improve their performance on our task. However, given their off-the-shelf adoption for downstream tasks like automatic commentary generation, these models are expected to possess the prerequisite skill of recognizing important moments, which is what we test. For the proprietary model, we switch off grounding with online web search so that it cannot retrieve the highlight reels used to construct our dataset.

Finally, although our logit-based analysis is purely observational and cannot causally prove modality collapse in models, it clearly highlights the limitations of current architectures. Demonstrating this relationship causally between multimodal integration and model confidence remains an open question for future work.

\section*{Acknowledgments}
We are grateful to our colleagues at the Dialogue Modelling Group (DMG) for their insightful feedback at different stages of this work.

\bibliography{custom}

\appendix

\section{Constructing the MOMENTS Dataset}
\label{appendix:A}
\begin{figure}[htbp]
  \centering
  \includegraphics[width=\columnwidth,keepaspectratio]{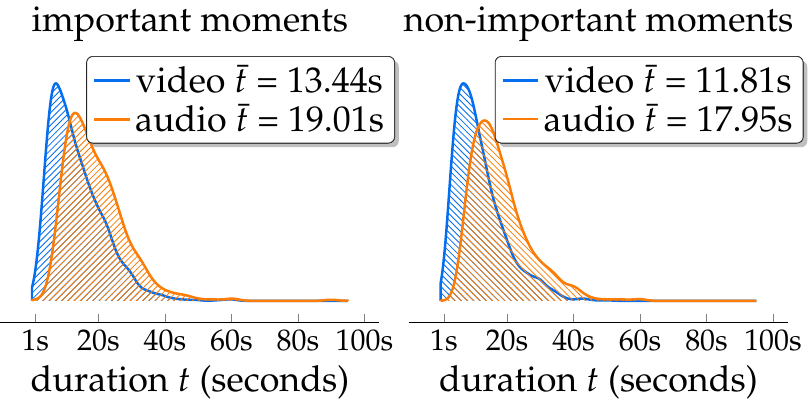}
  \caption{The duration distributions of non-important moments (NIM) reflect the distributions of important ones (IM), with the average duration ($\bar{t}$) of the audio modality extending beyond the average of corresponding video frames for both classes.}
  \label{fig:moments_duration}
\end{figure}

\begin{figure}[tb]
  \centering
  \includegraphics[width=\columnwidth,keepaspectratio]{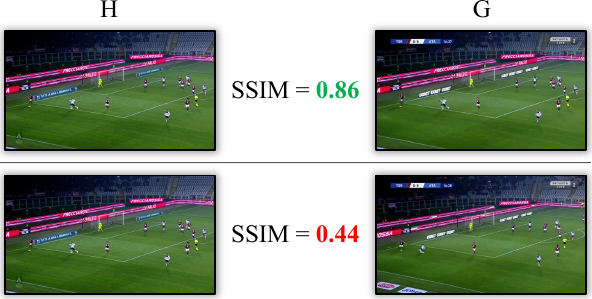}
  \caption{On the top, we compare frames from an example <H,G> pair both pertaining to the exact same instance of the game (accounting for frame rates). Differences in terms of scorecards and advertisement overlays are penalized appropriately by the SSIM metric (0.86); On the bottom, the frame from G pertains to an instance in the same game that is half a second apart from the instance depicted by the H frame. In this case, SSIM (0.44) accurately reflects all valid visual differences in terms of player formations and ball position.}
  \label{fig:SSIM}
\end{figure}

\begin{figure}[tb]
  \centering
  \includegraphics[width=\columnwidth,keepaspectratio]{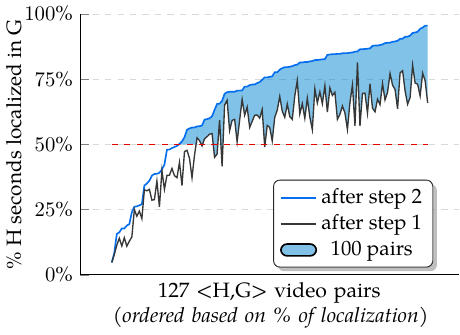}
  \caption{Among the 127 <$\text{H}$,$\text{G}$> pairs we process, more than half of the frames in 100 highlight ($\text{H}$) videos are accurately localized in their corresponding full-game ($\text{G}$) videos after step 2.}
  \label{fig:after_steps_1_2}
\end{figure}

Figure~\ref{fig:SSIM} illustrates the behavior of the Structural Similarity Index Measure (SSIM) when comparing candidate frames between the highlight reels (H) and the full game videos (G). The metric effectively assigns high scores (e.g., SSIM = 0.86) to true visual matches despite auxiliary differences (scorecards, advertisement overlays), while appropriately reflecting truly different frames (e.g., SSIM = 0.44 for visuals that are just half-second apart). Figure~\ref{fig:after_steps_1_2} plots the impact of our search space pruning technique, showing a substantial increase in accurately matched frames after step 2. Figure~\ref{fig:moments_duration} compares the duration distributions of the extracted important and non-important moments. We fit a Gamma distribution rather than other standard distributions (e.g., Normal or Exponential) given its flexibility and the properties of moment durations, which are non-negative and tend to be right-skewed considering the editing process. Furthermore, to verify the integrity of this automated extraction, we conduct a human sanity check on small subsets of moments randomly drawn from 10 football games.

\section{Experiments}
\label{appendix:B}
We use the HuggingFace transformers library to access pre-trained model weights and utilize Flash Attention \cite{FlashAttention} for optimizing memory and inference compute times. Prompts used for obtaining responses from models are provided below:
\begin{table}[!b]
  \centering
  \begin{tabularx}{\columnwidth}{X}
    \hline
    \cellcolor{xkcdPaleMauve!60}\textbf{System Prompt:} You are an expert of soccer games, who is well versed with the knowledge of the sport, and capable of recognizing game highlights.\\\\
    \cellcolor{xkcdCreme!80}\textbf{Generation Prompt 1:} Classify whether this is an important moment of the game that should be included in the highlights. Here is the commentary associated with this moment: \small``\{<\texttt{COMMENTARY}>\}''\normalsize. Respond only with a ``YES'' or a ``NO''.\cellcolor{xkcdCreme!80}\\
    \cellcolor{xkcdCreme!80}\textbf{Generation Prompt 2:} Is this an important moment of the game that is worthy of inclusion in the highlights reel? This is the transcription of the commentary: \small``\{<\texttt{COMMENTARY}>\}''\normalsize. Respond only with a ``YES'' or a ``NO''.\\
    \hline
  \end{tabularx}
  \caption{System prompt and paraphrased generation prompts used to evaluate all the models. When the commentary is not among the input modalities, the sentence enclosing it is omitted.}
  \label{tab:prompts}
\end{table}

\begin{figure*}[t]
  \centering
  \includegraphics[width=0.97\textwidth]{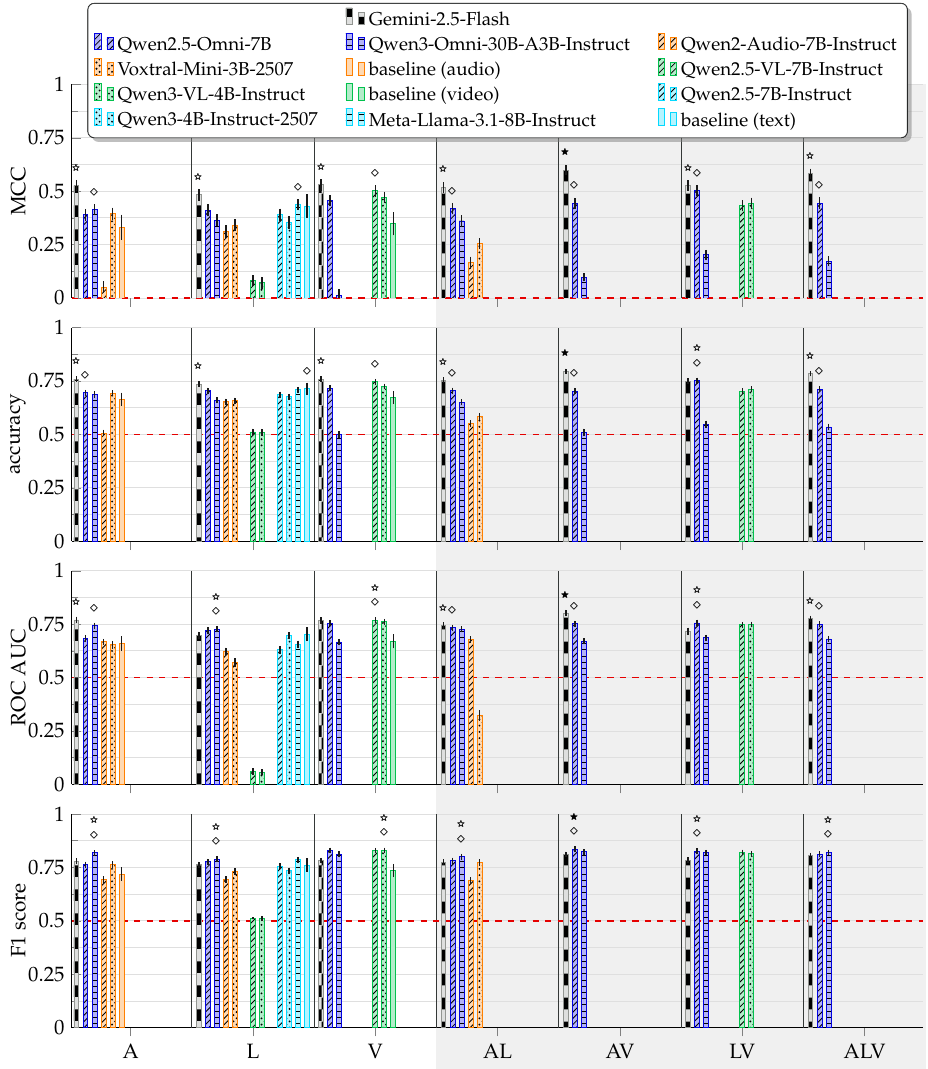}
  \caption{MCC, accuracy, ROC AUC, and F1 scores for all models under various combinations of modalities. Under each modality combination, $\smallstar$ denotes the highest scoring model and $\diamond$ denotes the highest scoring open source model. $\filledstar$ represents the overall best performing model and combination of modalities.}
  \label{fig:scores_supplementary}
\end{figure*}

We evaluate model responses using four standard metrics used for classification tasks---MCC, accuracy, ROC AUC, and F1 score. Figure~\ref{fig:scores_supplementary} visualizes the scores from these metrics for all the models prompted under different modality settings. Consistent with our observations in terms of MCC and accuracy, scores obtained using the ROC AUC and F1 metrics reflect that performance across models is largely comparable. In addition, Table~\ref{tab:confusion_matrices} provides the full confusion matrices for three representative high-scoring models across different modality settings.

\newcolumntype{C}{>{\centering\arraybackslash}m{1.8cm}}
\newcommand{\cmat}[4]{
    \pgfmathsetmacro{\total}{#1+#2+#3+#4}
    \pgfmathtruncatemacro{\pa}{round((#1/\total)*100)}
    \pgfmathtruncatemacro{\pb}{round((#2/\total)*100)}
    \pgfmathtruncatemacro{\pc}{round((#3/\total)*100)}
    \pgfmathtruncatemacro{\pd}{round((#4/\total)*100)}

    \pgfmathtruncatemacro{\pbcol}{round((#2/\total)*50)}
    \pgfmathtruncatemacro{\pccol}{round((#3/\total)*50)}
    \begin{tabular}{|c|c|}
        \hline
        \cellcolor{white} \pa & \cellcolor{red!\pbcol} \pb \\ \hline
        \cellcolor{red!\pccol} \pc & \cellcolor{white} \pd \\ \hline
    \end{tabular}
}

\begin{table*}[tbp]
\centering
\resizebox{\textwidth}{!}{

\begin{tabular}{m{3.5cm} C C C C C C C}
    & \textbf{A} & \textbf{L} & \textbf{V} & \textbf{AL} & \textbf{AV} & \textbf{LV} & \textbf{ALV} \\[1ex]
    \textbf{Gemini-2.5-Flash}  & \cmat{1673}{304}{650}{1327} & \cmat{1681}{296}{751}{1226} & \cmat{1708}{269}{684}{1293} & \cmat{1673}{304}{668}{1309} & \cmat{1728}{249}{559}{1418} & \cmat{1811}{166}{832}{1145} & \cmat{1769}{208}{650}{1327} \\[4ex]
    \textbf{Qwen2.5-Omni-7B}   & \cmat{1303}{674}{536}{1441} & \cmat{1505}{472}{699}{1278} & \cmat{1714}{263}{857}{1120} & \cmat{1615}{362}{810}{1167} & \cmat{1792}{185}{993}{984} & \cmat{1524}{453}{532}{1445} & \cmat{1721}{256}{889}{1088} \\[4ex]
    \textbf{Qwen2.5-VL-7B-Instruct} & & \cmat{62}{1915}{17}{1960} & \cmat{1661}{316}{687}{1290} & & & \cmat{1745}{232}{951}{1026} & \\[6ex]
\end{tabular}
}
\caption{Confusion matrices with percentages for three high-scoring models. In each matrix, clockwise from the top left are values for: \textit{True Important}, \textit{False Non-important}, \textit{False Important}, and \textit{True Non-important} predictions. The off-diagonal of matrices represent error percentages with a stronger shade of red indicating a higher percentage. We observe that models struggle more with non-important compared to important moments.}
\label{tab:confusion_matrices}
\end{table*}

\section{Analyses}
\label{appendix:C}
\subsection{Influence of Modalities}
Figure~\ref{fig:contribution_scores_supplementary} shows modality-level contribution scores for all the models we considered in this study. For multimodal models capable of processing videos, we largely notice that the visual modality contributes the most to models' confidence for correctly identifying important moments, whereas the associated commentaries are crucial in determining the confidence of models for correctly classifying non-important moments. For audio-language and unimodal models, we do not observe any modality contribution patterns that consistently correlate with importance of moments.

\begin{theorem}
  The difference in logits between two classes (tokens) equals the difference in their $\log$ probabilities.
\end{theorem}

\begin{proof}
  Using the definition of softmax, probability of a token $i$,
  \begin{equation*}
    p_i = \frac{e^{z_i}}{\sum_{j \in V} e^{z_j}},
  \end{equation*}
  where $z_i$: logit for token $i$ and $V$: the entire vocabulary. Based on this, the $\log$ probability of token $i$ =
  \begin{equation*}
    \log p_i = z_i - \log \left( \sum_{j \in V} e^{z_j} \right).
  \end{equation*}
  Thus, for any two classes (e.g., YES and NO), the difference in their log probabilities
  \begin{align*}
    \log p_{\text{YES}} - \log p_{\text{NO}}
    &= \left( z_{\text{YES}} - \log \sum_{j \in V} e^{z_j} \right) \\
    &\quad - \left( z_{\text{NO}} - \log \sum_{j \in V} e^{z_j} \right)
  \end{align*}
  equals the difference in their logits upon canceling the normalization terms:
  \begin{equation*}
    \log p_{\text{YES}} - \log p_{\text{NO}} = z_{\text{YES}} - z_{\text{NO}}. \qedhere
  \end{equation*}
\end{proof}

\subsection{Role of Multimodality}
To obtain a set of prototypical (goal) and contextual (shot-on-target, corner kick/throw-in) moments for our analyses, we followed a two-stage annotation process. First, for each type (e.g., goal), we prompt a foundation model to automatically identify moments in our dataset that encapsulate the action. Specifically, for each moment, we obtain confidence of model ($\Delta{Z}$) under the three modalities in scope, separately. Prompts used are provided below, and we obtain the \small``\{<\texttt{DEFINITION}>\}''\normalsize~for each moment type---goals, corner kicks, shot-on-target, throw-in---from \citet{action_definitions}.
\begin{center}
  \begin{tabularx}{\columnwidth}{X}
    \hline
    \cellcolor{xkcdPaleMauve!60}\textbf{System Prompt:} You are an expert of soccer games, who is well versed with the knowledge of the sport, and capable of identifying \small``\{<\texttt{TYPE}>\}''\normalsize~moments.\\\\
    \cellcolor{xkcdCreme!80}\textbf{Generation Prompt:} \newline{(with text)} Is this commentary: \small``\{<\texttt{COMMENTARY}>\}''\normalsize; {(with video)} Is this video clip; {(with audio)} Is this audio commentary about a \small``\{<\texttt{TYPE}>\}''\normalsize~moment? A \small``\{<\texttt{TYPE}>\}''\normalsize~is $\dots$ \small``\{<\texttt{DEFINITION}>\}''\normalsize. Respond only with a ``YES'' or a ``NO''.\\
    \hline
  \end{tabularx}
\end{center}

\begin{figure*}[tbp]
  \centering
  \includegraphics[width=\textwidth]{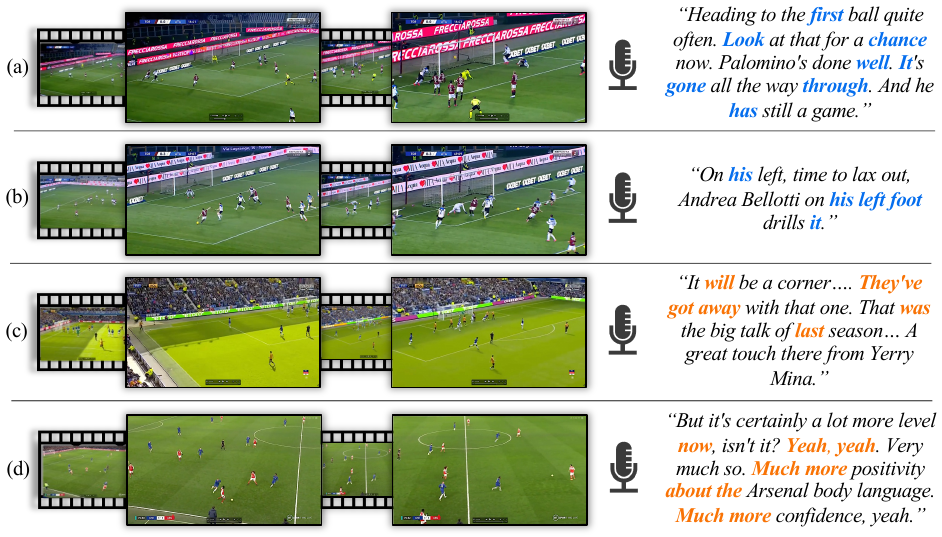}
  \caption{Examples from our MOMENTS dataset: (a) shows a `goal' scoring moment, (b) shows a `shot-on-target' saved by a goalkeeper from the opposing team, (c) shows a non-important `corner kick' that did not result in a goal, and (d) shows another non-important moment in which players are passing the ball in the midfield.}
  \label{fig:moments_examples_supplementary}
\end{figure*}

Then, for each type, we select moments that are categorized reliably---$\Delta{Z} > 3.0$ \cite{logit_difference_range}---as ``YES'' under all the three modalities. Finally, we verify the selected moments manually to ensure correctness of type and completeness of the action. Examples of selected moments are provided in Figure~\ref{fig:moments_examples_supplementary}.

\section{Reproducibility}
\label{appendix:D}
All the details for reproducing the experiments and analyses are included in the accompanying code. We provide all hyperparameter configurations and random seeds utilized to obtain the reported results. Furthermore, we conducted our evaluations using a single H100 GPU and list the exact versions of all required software libraries in the \texttt{requirements.txt} file.

\begin{figure*}[tbp]
  \centering
  \hfill
  \begin{subfigure}{0.47\textwidth}
    \includegraphics[width=\linewidth]{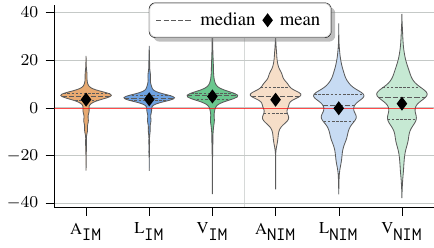}
    \caption{Gemini-2.5-Flash}
  \end{subfigure}
  \hfill
  \vspace{1em}

  \begin{subfigure}{0.47\textwidth}
    \includegraphics[width=\linewidth]{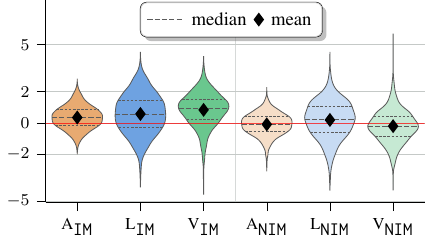}
    \caption{Qwen2.5-Omni-7B}
  \end{subfigure}
  \hfill
  \begin{subfigure}{0.47\textwidth}
    \includegraphics[width=\linewidth]{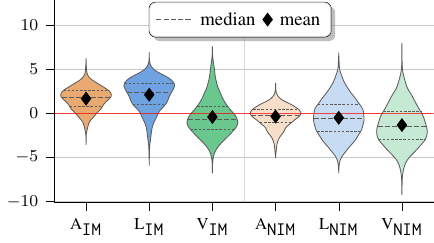}
    \caption{Qwen3-Omni-30B-A3B-Instruct}
  \end{subfigure}

  \vspace{1em}

  \begin{subfigure}{0.47\textwidth}
    \includegraphics[width=\linewidth]{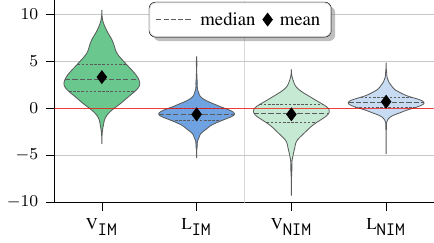}
    \caption{Qwen2.5-VL-7B-Instruct}
  \end{subfigure}
  \hfill
  \begin{subfigure}{0.47\textwidth}
    \includegraphics[width=\linewidth]{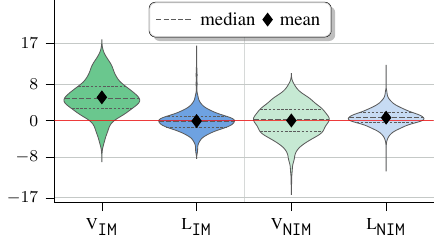}
    \caption{Qwen3-VL-4B-Instruct}
  \end{subfigure}

  \vspace{1em}

  \begin{subfigure}{0.47\textwidth}
    \includegraphics[width=\linewidth]{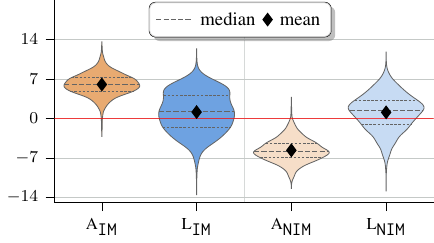}
    \caption{Qwen2-Audio-7B-Instruct}
  \end{subfigure}
  \hfill
  \begin{subfigure}{0.47\textwidth}
    \includegraphics[width=\linewidth]{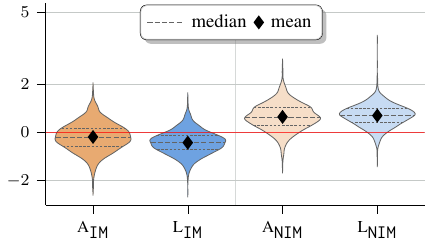}
    \caption{Voxtral-Mini-3B-2507}
  \end{subfigure}
\end{figure*}

\begin{figure*}[tbp]
  \ContinuedFloat
  \centering
  \begin{subfigure}{0.47\textwidth}
    \includegraphics[width=\linewidth]{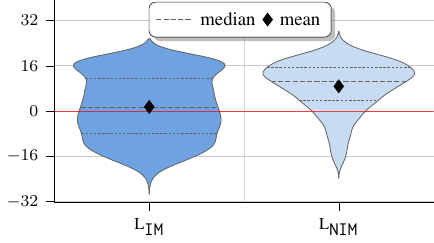}
    \caption{Qwen2.5-7B-Instruct}
  \end{subfigure}
  \hfill
  \begin{subfigure}{0.47\textwidth}
    \includegraphics[width=\linewidth]{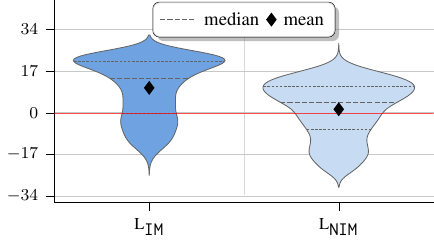}
    \caption{Qwen3-4B-Instruct}
  \end{subfigure}
  \hfill
  \begin{subfigure}{0.47\textwidth}
    \includegraphics[width=\linewidth]{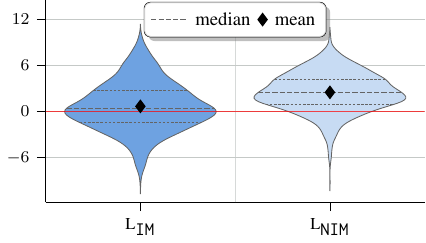}
    \caption{Meta-Llama-3.1-8B-Instruct}
  \end{subfigure}
  \hfill
  \begin{subfigure}{0.47\textwidth}
    \includegraphics[width=\linewidth]{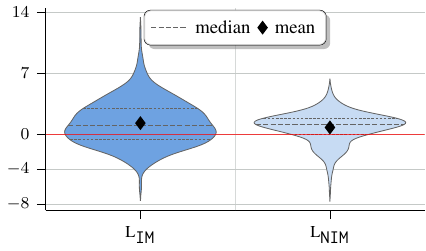}
    \caption{baseline (text)}
  \end{subfigure}

  \vspace{1em}

  \begin{subfigure}{0.47\textwidth}
    \includegraphics[width=\linewidth]{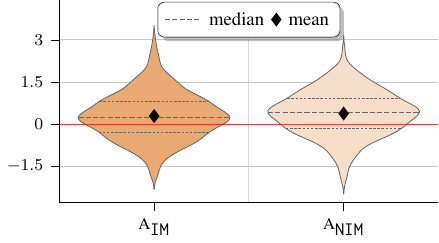}
    \caption{baseline (audio)}
  \end{subfigure}
  \hfill
  \begin{subfigure}{0.47\textwidth}
    \includegraphics[width=\linewidth]{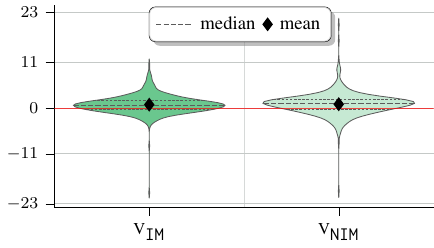}
    \caption{baseline (video)}
  \end{subfigure}
  \caption{Modality-level contribution scores for all the models we considered in this study, separated for important (\texttt{IM}) and non-important (\texttt{NIM}) moments.}
  \label{fig:contribution_scores_supplementary}
\end{figure*}

\end{document}